\documentclass{article}
\usepackage{graphicx} 
\usepackage{vilyastyle}
\usepackage{hyperref}
\usepackage[backend=biber,style=numeric, natbib=true, sorting=none]{biblatex}
\title{Vilya-1: An all-atom foundation model for macrocycle structure prediction and design}

\begin{document}

\maketitle
\thispagestyle{titlepage}

\begin{abstract}
Macrocyclic peptides are an increasingly important therapeutic modality, but existing computational methods for modeling their structures and properties are limited in scope and do not generalize well across the synthetically accessible chemical space. In this work, we introduce Vilya-1, a deep learning model that addresses two central challenges in macrocycle design: sampling biologically relevant conformations across arbitrary chemistries and predicting key developability properties such as membrane permeability. Vilya-1 operates on a uniform all-atom representation and is trained on heterogeneous structural datasets spanning diverse topologies and chemical classes. Across a broad set of macrocycles composed of canonical and non-canonical residues, Vilya-1 substantially improves geometric accuracy relative to physics-based methods, co-folding networks, and deep-learning conformer generators, while maintaining broad chemical coverage that extends to small molecules. Vilya-1 also supports generative applications, enabling the design of novel macrocycles with tailored chemical, structural, and property profiles. Together, these capabilities establish Vilya-1 as a foundation model for accelerating the development of next-generation macrocycle therapeutics.
\end{abstract}

\section{Introduction}
Macrocyclic peptide therapeutics offer the potential to bind to traditionally undruggable protein surfaces while retaining many of the pharmacological advantages of small molecules, particularly the ability to access intracellular compartments and the potential for oral dosing \cite{garcia2023macrocycles, viarengo2025macrocycles}. \textit{De novo} design studies have shown that macrocycles composed of canonical amino acids can be engineered with atomic level accuracy, vastly expanding the set of scaffolds beyond those found in nature \cite{bhardwaj2016accurate, rettie2025accurate, watson2023novo}. More recently, computational pipelines have been extended to chemically diverse macrocyclic backbones and non-peptidic macrocyclic oligoamides \cite{salveson2024expansive}. This chemical diversity opens the possibility for macrocycles to address challenging targets involving both intra and extracellular protein-protein interactions. These advances have highlighted the importance of accurately modeling the low-energy conformational states a macrocycle may adopt, as these states control target binding, permeability, and other drug-like properties. To date, however, conformational sampling has generally relied on custom computational procedures tailored to specific chemistries. In order to unlock the therapeutic potential of macrocycles, it is necessary to develop computational methods that generalize across synthetically accessible chemistries and accurately capture biologically relevant low-energy states.

\begin{figure}[htbp]
  \centering
  \includegraphics[width=1.0\linewidth]{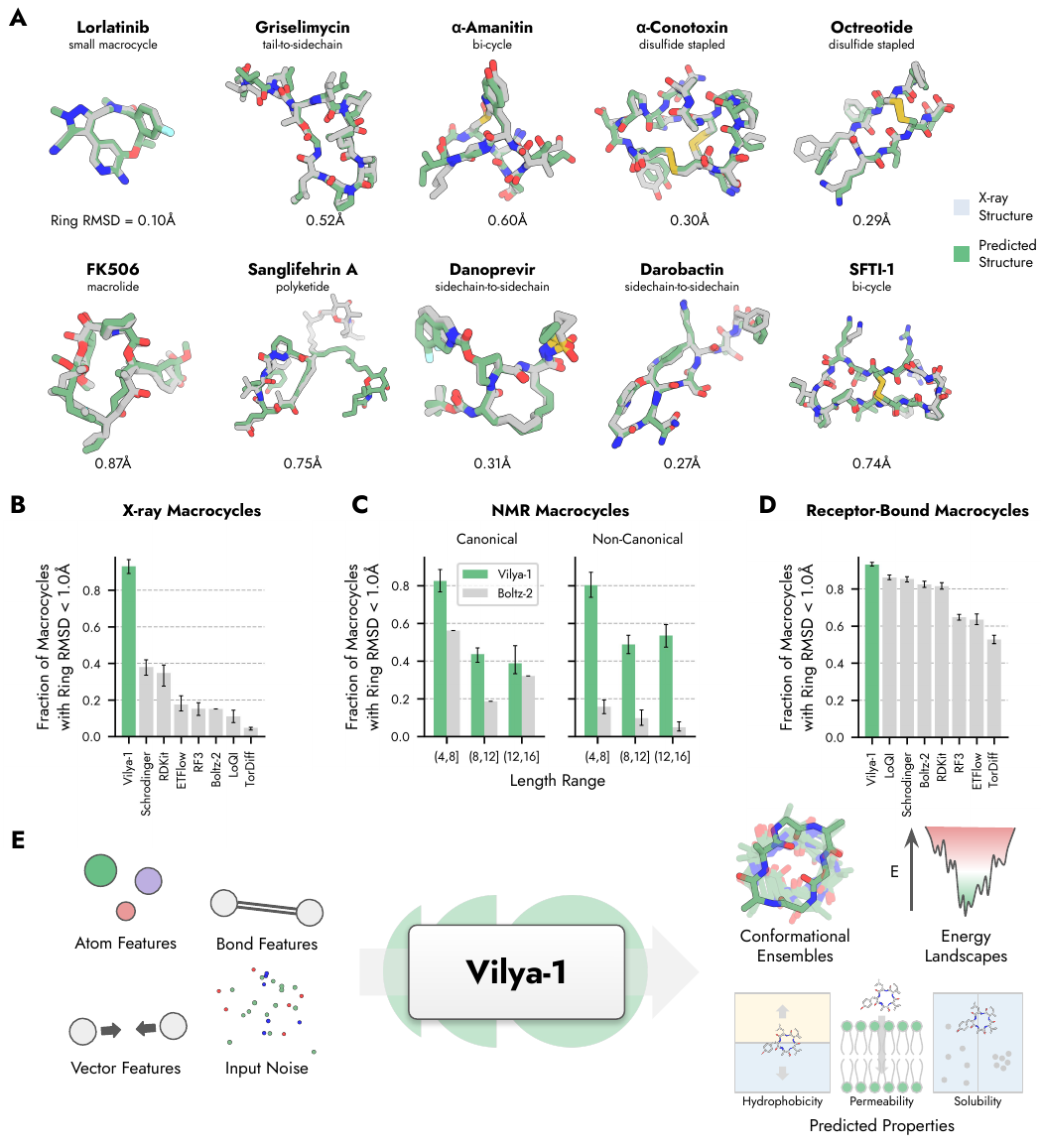}
  \caption{Bioactive conformer prediction and sampling efficiency of Vilya-1. A) Examples of bioactive macrocycles for which Vilya-1 predicts conformations within sub-angstrom accuracy, measured by ring RMSD. Vilya-1-generated structures (green) are overlaid with the corresponding experimental X-ray structures from the PDB (gray). B) Sampling efficiency of Vilya-1 on X-ray structures of cyclic peptides, compared with existing methods. C) Performance on solution NMR structures of canonical and non-canonical cyclic peptides. D) Performance on receptor-bound conformations of cyclic peptides, showing higher success rates for Vilya-1 relative to alternative methods. E) Schematic overview of Vilya-1. For panels A--D, 100 conformers are generated per molecule with each method including Vilya-1. For panel A, the generated conformers are also ranked by confidence and the best ring RMSD model among the top 5 is reported, mimicking the CASP competition setting \cite{moult1995large}. Results shown in panels B--D reflect conformational sampling performance only, with no scoring or confidence-based ranking. }
  \label{fig:fig1}
\end{figure}

Existing methods to predict the structures of macrocyclic peptides largely derive from tools for protein structure prediction or small molecule conformer generation. Both have important limitations for chemically-diverse cyclic systems. Generalized structure prediction networks such as Boltz-2 and RosettaFold3 (RF3) \cite{passaro2025boltz, corley2025accelerating} can in principle model macrocycles, yet in practice do not generalize well to non-canonical chemistries and typically fail to produce sufficiently diverse conformational ensembles for macrocycles that may lack a single, well-defined ground state. Although there are examples of fine-tuning co-folding networks on structural data for macrocyclic peptides containing non-canonical amino acids, these approaches have not demonstrated robust generalization beyond the narrow chemistries represented in their training sets \cite{li2025rarefold, cao2025accurate, zhu2025predicting, mao2025ncpepfold}. Small-molecule conformer generators are designed for diversity but often do not generalize to larger, flexible molecules with ring constraints. Physics-based protocols that use molecular dynamics or stochastic sampling are capable of modeling highly complex macrocycles, but they are inefficient and their accuracy tends to decrease on larger macrocycles \cite{sindhikara2017improving,jain2019complex,damjanovic2021elucidating}. This gap motivates the development of new methods specifically tailored to model macrocycle conformational landscapes.

Here we introduce Vilya-1, a deep learning model for generating conformers of macrocycles. Vilya-1 operates on a uniform all-atom representation that does not distinguish between peptide and small-molecule chemistries, enabling a single architecture to accommodate the broad range of chemistries encountered in macrocycle design. This representation simplifies the use of molecular structural datasets and lets the model learn generic structural principles governing both polymer and non-polymer molecules. Vilya-1 is trained on an extensive set of internal and external structural datasets spanning diverse topologies, chemical classes and molecular sizes. Across a broad suite of benchmarks, Vilya-1 substantially outperforms existing approaches in sampling experimentally observed macrocycle structures, nearly doubling the success rate of accurate ring reconstruction (ring RMSD $< 1 $\AA) relative to the leading physics-based methods and markedly exceeding the performance of co-folding networks, particularly for molecules containing non-canonical amino acids or non-peptidic fragments.

We show the utility of accurate structure sampling on three real-world applications: (i) as a foundation model for downstream property prediction, (ii) as a sampler for conformational landscape analysis, and (iii) as an oracle for designing novel topologies. As a property prediction model, Vilya-1 is capable of learning from internal assay data to predict key drug-like properties that enable oral bioavailability. Conformational landscapes generated by Vilya-1 are predictive of binding to protein targets, despite the model not being trained for binder discrimination. Finally, as part of a generative pipeline, Vilya-1 is capable of scaffolding motifs from larger macrocycles into smaller ones, an important problem in hit optimization of drug-like macrocycles. Taken together, these results demonstrate that Vilya-1 is a general-purpose model capable of aiding in all stages of designing and optimizing macrocyclic peptides in real-world drug discovery campaigns.

\section{Methods}
\subsection{Vilya-1 Architecture}

The architecture is inspired by existing work on scaling large transformer models to biomolecular modeling problems \cite{abramson2024accurate, anishchenko2025modeling, geffner2025proteina}. As input, the model receives a complete chemical description of a molecule in three different modalities: (i) atom-level scalar features, (ii) pairwise scalar features, and (iii) atom-level vector features. Aided by specialized NVIDIA cuEquivariance kernels, the core architecture is built on top of components introduced by \citet{jumper2021highly}: triangle attention and multiplication, and attention with pair bias. The model then projects the final embeddings into coordinates directly in 3D space.

\begin{figure}[htbp]
  \centering
  \includegraphics[width=1.0\linewidth]{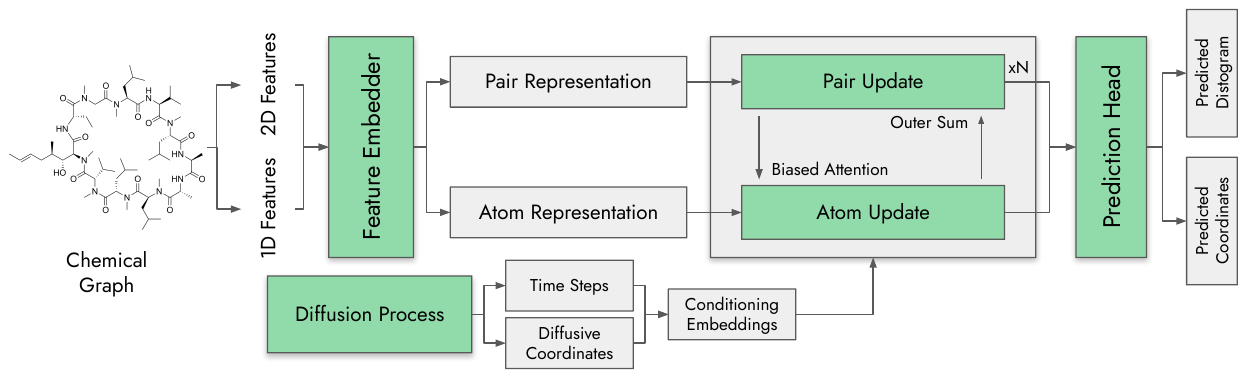}
  \caption{Schematic of Vilya-1 architecture. Model architecture and noise process components are indicated in green, while feature embeddings and inputs are indicated in gray.}
  \label{fig:fig2}
\end{figure}

The model differs from popular protein ligand co-folding architectures in three main ways. First, the diffusion process runs through a single, unified transformer architecture. As our goal is to explicitly model an ensemble of conformations, we hypothesize that having separate trunk and diffusion modules may reduce the diversity of sampled coordinate structures. Second, because we aim to model a broad diversity of non-canonical chemistries, we embed all molecular inputs at the level of individual heavy atoms. While explicitly grouping atoms in canonical residues together in a single embedding is computationally efficient, it limits the generality of the network. Finally, we only input features that describe the chemical structure of the input molecule and omit auxiliary features like molecule type, atom name, or positional encoding. We reason that having only chemically relevant features will make the model less likely to memorize training data and more likely to generalize to novel molecules.

We extend Vilya-1 architecture with two downstream models: a confidence model and a property prediction model. Both downstream models share the architecture of the conformer generation model and are fine-tuned from its weights. Both property and confidence models take in the same features as the base model, in addition to the structure of a generated conformer.

\subsection{Training}
\paragraph{Conformer generator:} The model is trained on a combination of publicly-available small molecule and peptide crystal structure data in addition to computationally-generated structures of macrocyclic peptides. The primary loss is a diffusion-based loss to reconstruct the coordinates given the chemical input features, current noise level, and noisy coordinates. We find it useful to have two auxiliary losses: a pairwise distogram loss similar to \citet{abramson2024accurate}, and an additional loss that encourages the correct chirality of tetrahedral centers.

\paragraph{Confidence estimation:} The confidence model is trained to predict a modified version of Frame Aligned Point Error (FAPE) between a generated conformer and the true structure \cite{krishna2024generalized, anishchenko2025modeling}.

\paragraph{Property prediction:} The property prediction model is trained to jointly predict multiple physicochemical properties from generated conformers. Predicted labels include internally measured properties\textemdash permeability, hydrophobicity measured as chromatographic LogD, and kinetic solubility\textemdash as well as a computed property, 3D polar surface area, derived directly from the input conformer. Prediction tasks are formulated as either regression or sigmoid regression depending on the nature of each label.

Training and evaluation data for property prediction is aggregated from two sources: (i) internal datasets collected during Vilya’s discovery programs and (ii) externally available permeability datasets compiled from published assay measurements. Passive permeability is modeled primarily using data from parallel artificial membrane permeability assays (PAMPA). The external PAMPA dataset comprises approximately 11,500 unique measurements collected from the Cyclic Peptide Membrane Permeability Database \cite{li2023cycpeptmpdb}, the Non-Peptidic Macrocycle Membrane Permeability Database \cite{feng2024development}, and other databases and published studies \cite{juravcka2019molmedb, huang2019ncats, lomize2019permm, rzepiela2022conformational}. Additional auxiliary prediction tasks are trained exclusively on proprietary data, generated at Vilya, which includes permeability measured in the Madin-Darby canine kidney (MDCK) assay, chromatographic LogD, and kinetic solubility. The computed property, 3D polar surface area, is computed on the fly from the predicted conformer and serves as an auxiliary target to promote consistent coupling between molecular geometry and predicted properties. Data is split using time-based splitting for internal data mimicking real world usage of property prediction \cite{landrum2023simpd}, and scaffold-based splitting for external data.

\subsection{Evaluation Sets for Macrocycle Structure Prediction}
\paragraph{Small molecule validation set:} We assembled a validation set of 7,192 organic small-molecule structures from the Cambridge Structural Database (CSD), spanning 3--96 heavy atoms \cite{groom2016cambridge}. These experimentally determined geometries are used to benchmark performance on chemically diverse, non-peptidic systems. 

\paragraph{Macrocycle validation set:} We curated a set of 200 amide-containing macrocycles mined from the CSD with $\geq$12 atoms and at least three amides in the backbone ring. The CSD identifiers for all selected structures are provided in the Supplement. This set serves as the primary resource for hyperparameter tuning and model selection.

\paragraph{Macrocycle test set of X-ray structures:} We compiled a test set of 66 cyclic peptides with X-ray structures from three sources: 53 examples from recent literature \cite{bhardwaj2022accurate, salveson2024expansive, rettie2025cyclic}, 5 internal ligand-only structures, and 8 additional macrocycles from the Protein Data Bank (PDB). These peptides range in size from 3 to 17 amino-acid residues.

\paragraph{Macrocycle test set of NMR structures:} We established an independent benchmark of 260 NMR-derived structures constructed from entries curated by the Cyclic Peptide DataBank with $\leq$16 residues, at least one macrocyclic ring system, and NMR-determined coordinates. Structures were sanitized and standardized to ensure the absence of radical electrons and consistency in valency, formal charge, and hydrogen counts using a pipeline based on Biotite and RDKit \cite{kunzmann2018biotite, greg_landrum_2025_18098214}. The final set was further divided into all-canonical and non-canonical subsets containing 86 and 174 molecules respectively. PDB IDs of the selected molecules are provided in the Supplement.

\paragraph{PDB-derived bound structures of small molecules:} To benchmark performance on ligand-protein complexes, we use the Platinum Diverse dataset, a high-quality collection of 2,859 protein-bound ligand conformations extracted from the PDB and designed specifically to evaluate conformer ensemble generators \cite{friedrich2017high}.

\paragraph{PDB-derived bound structures of macrocycles:} We additionally evaluate our method on a recently published benchmark of 240 macrocyclic ligands extracted from receptor-bound structures in the PDB. Ligands in this set are classified as small-molecule macrocycles or cyclized peptides, and all contain $\leq$40 ring heavy atoms. While the ligands in the data set were prepared by the authors in \citet{robson2025accurate}, we further corrected bond orders and protonation states to better reflect the likely binding state.

Molecules which share more than 70\% Tanimoto similarity (Morgan circular fingerprints 2048) to any of the test macrocycles were excluded from training.

\subsection{Evaluation Methodology}
\paragraph{Conformer generation:} To evaluate conformer generation performance, we sample 100 conformers from each method and take the minimum RMSD between the predicted structures and the ground truth structure. For macrocycles, the RMSD is computed by aligning the predicted and ground truth structures on the ring scaffold atoms, and then computing the RMSD over those same atoms. We define the ring scaffold as the set of heavy atoms that comprises the central macrocyclic ring, all directly substituent heavy atoms, and any fused or linked ring systems connected to that central ring (Fig \ref{fig:figs1}). This definition of the ring scaffold captures all atoms whose coordinates are solely dependent on correctly predicting the locations of all heavy atoms that comprise the central ring and the bond torsions between them. As such, it punishes large torsional deviations in the central ring, such as $180^\circ$ rotations of amide bonds, that can be masked by more simple C$\alpha$-only, or main chain-only superpositions which are commonly used in the literature. For macrocycles, we consider a prediction a success if its ring scaffold-based RMSD (ring RMSD) is $< 1$\AA. For small molecules, we consider a prediction a success if its all-atom RMSD is $< 0.5$\AA. A method is considered to successfully predict a crystal structure if any individual prediction from the generated ensemble is classified as a success. For NMR structures, we compute ring RMSD to every member of the ensemble and pick the minimum RMSD across the ensemble. Results for all-atom RMSD on macrocycles are in Fig \ref{fig:figs7}.

The structure prediction capabilities of Vilya-1 were benchmarked against three different classes of methods. For classical methods, we benchmark against Schrodinger’s Prime-MCS (\verb|treat_bad_torsions fix, spinroot 10|) and RDKit’s ETKDGv3 with macrocyclic torsion profiles and 1,4 distance bounds \cite{sindhikara2017improving, wang2020improving, robson2025accurate}. In the class of deep-learning based conformer generators, we benchmark against LoQI, ETFlow and TorDIFF \cite{nikitin2025scalable, jing2022torsional, hassan2024flow}. We omit a recent method, RINGER, as it is unable to model the majority of compounds in our test set due to the presence of non-canonical chemistries, but include a head-to-head comparison on canonical macrocycles in Fig \ref{fig:figs2} \cite{grambow2024accurate}. For co-folding methods, we benchmark against Boltz-2 and RF3, and omit other methods due to commercial license restrictions \cite{passaro2025boltz, corley2025accelerating}.

\paragraph{Confidence estimation:} To evaluate our confidence model, we compare its ability to rank order high-accuracy conformers to the ranking output by recent machine learning interatomic potentials (MLIP), as well as random selection \cite{fu2025learning}. We generate 100 conformers for each molecule, take the top prediction from each ranking method, and then compute the minimum ring RMSD to that prediction. We then compute the success rate as the number of molecules for which the top-ranked prediction is below the $< 1$\AA\ ring RMSD threshold.

\paragraph{Property prediction:} To evaluate our property prediction model, we compare its performance to ChemProp and its pretrained version CheMeleon, which are graph-based models commonly used in property prediction \cite{burns2025descriptor, heid2023chemprop, burns_2025_15426601}. We use a simple K-Nearest Neighbor (KNN) model fit on Morgan fingerprints of the compounds as the baseline model. An ablation study is performed to compare the performance of these architectures under the effect of data (internal vs. external) and initialization (scratch vs. pretrained), where applicable. Model performance was evaluated using the enrichment factor (EF), defined as the fold enrichment of desirable compounds among the top 10\% of model-ranked candidates relative to their background frequency. Higher EF values indicate more effective prioritization. The success thresholds are $0.5 \times 10^{-6}$ cm/s for PAMPA and MDCK permeability and 60 $\mu$M for Kinetic Solubility.

\section{Results}
\subsection{Vilya-1 is the best-in-class conformer sampler for macrocycles}
\paragraph{Vilya-1 recapitulates X-ray and solution NMR structures.} Representative examples in Fig \ref{fig:fig1}A  drawn from the PDB illustrate the ability of Vilya-1 to predict biologically relevant conformations of bioactive macrocycles across diverse chemical classes and topologies. Remarkably, Vilya-1 achieves sub-angstrom accuracy in all 10 examples shown, despite the fact that, for this analysis, no PDB-derived structures were used during training, underscoring the model’s ability to generalize to entirely unseen experimental data. These examples include both non-peptidic macrolide (FK506) and polyketide (Sanglifehrin A) natural products, small constrained macrocycles (Lorlatinib), head-to-tail cyclized peptide macrocycles that contain side-chain crosslinks ($\alpha$-Amanitin, SFT1-1), large disulfide stapled peptides ($\alpha$-Conotoxin, Octreotide), sidechain-to-sidechain cyclized peptides (Danoprevir, Darobactin), as well as tail-to-sidechain cyclized peptides (Griselimycin). PDB IDs of the examples shown in Fig \ref{fig:fig1}A are listed in the Table \ref{tab:tabs1}.

On a benchmark comprising 66 X-ray structures of cyclic peptides, Vilya-1 samples a near-native ring conformation (ring RMSD $< 1$\AA) in 89.2\% of cases, more than doubling the success rate of the physics- and knowledge-based methods like Schrödinger Prime-MCS and RDKit ETKDGv3 (37.6\% and 34.5\%, respectively; Fig \ref{fig:fig1}B). The gap in performance is even bigger relative to co-folding methods (15.2\% for Boltz-2 and 14.8\% for RF3) and deep learning--based conformer generators (4.8\%, 10.6\%, 17.0\% for TorDiff, LoQI, and ETFlow respectively).

To assess the size limits of Vilya-1, we compiled a second benchmark of cyclic peptides with solution NMR structures deposited in the PDB, spanning sizes from 4 to 16 residues. Performance was evaluated separately for canonical and non-canonical peptides to probe the effect of residue- versus atom-based tokenization in Boltz-2. As shown in Fig \ref{fig:fig1}C, Boltz-2 exhibits substantial degradation in accuracy for non-canonical peptides, whereas Vilya-1 displays similar performance across canonical and non-canonical chemistries. Although the accuracy of all methods decreases as peptide size increases, Vilya-1 maintains a consistent advantage across the full range of sizes and chemical compositions examined. Success rates for other deep learning-based conformer generators on this benchmark were close to zero and were therefore excluded from the figure. Prime-MCS and ETKDGv3 exhibited long runtimes and high failure rates for molecules in this size range and were omitted from analysis for the same reason. RF3 showed qualitatively similar but consistently lower performance than Boltz-2 on this benchmark and was not included.

\paragraph{Vilya-1 samples receptor bound conformation of macrocycles and small molecules.} Vilya-1 is able to sample conformations of macrocycles that closely recapitulate experimentally observed bound structures, outperforming alternative approaches (Fig \ref{fig:fig1}D). To evaluate this capability, we used a recently released benchmark of 240 macrocyclic ligands derived from protein-ligand co-crystal structures in the PDB. Molecules in this dataset are considerably smaller than those in the preceding benchmarks, with a median ring scaffold size (Fig \ref{fig:figs1}) of 28 atoms, compared to median sizes of 47 and 54 atoms for the X-ray and NMR macrocycle test sets, respectively. In this regime, several existing methods\textemdash including RDKit, Boltz-2, Schrödinger, and LoQI\textemdash achieve success rates exceeding 80\%. We note that 202 of the 240 structures in this benchmark appear in the Boltz-2 training set, whereas Vilya-1 was trained on none of these structures. Despite the strong performance of these baselines, Vilya-1 attains the highest success rate of 93\%. A head-to-head comparison with Schrödinger’s Prime MCS further demonstrates Vilya-1’s advantage, with 92\% accuracy under a torsional ring RMSD threshold of $< 40^\circ$, compared to 90\% for Prime MCS (Fig \ref{fig:figs3}). Notably, Vilya-1 achieves this level of performance on bound macrocycle conformations without any task-specific training, highlighting its ability to generalize across both isolated and protein-bound conformational landscapes.

Although our goal was to develop a model that can accurately predict macrocycle structures, our approach is not specific to peptide chemistries. We evaluate our model on a standard test set of small molecules in their bound pose derived from the PDB that were not seen during training (Fig \ref{fig:figs5}). This test set measures the ability of methods to recapitulate biologically relevant structures of these small molecules. We find that Vilya-1 outperforms all other methods at small-molecule conformer generation, indicating that it is a general-purpose model that is applicable to a broad range of molecule types.

\begin{figure}[htbp]
  \centering
  \includegraphics[width=1.0\linewidth]{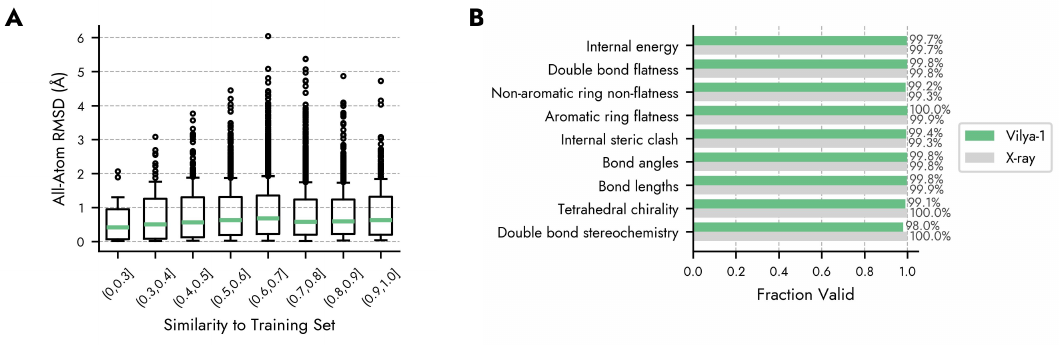}
  \caption{Generalization and chemical validity of Vilya-1 on a held-out set of small molecules. A) Vilya-1 accuracy as a function of Tanimoto similarity to the training set, showing no systematic dependence on similarity and indicating that performance is not driven by memorization. B) Chemical validity of Vilya-1-generated conformers assessed using PoseBusters, demonstrating validity rates comparable to reference X-ray structures deposited in the CSD. For both panels, evaluations were performed on a held-out validation set of 7,192 organic small molecules from the CSD.}
  \label{fig:fig3}
\end{figure}

\paragraph{Vilya-1 generalizes to molecules not seen during training.} A recurring critique of recent co-folding approaches is that their performance may be driven in part by memorization of the training set due to inadequate separation between training and test data, rather than by learning the underlying principles that govern biomolecular structure \cite{vskrinjar2025have, durairaj2024plinder, morehead2025deep}. To assess whether Vilya-1 exhibits similar behavior, we evaluated its performance on the small molecule validation set from the CSD. As shown in Fig \ref{fig:fig3}A, Vilya-1’s accuracy is effectively uncorrelated with a molecule’s Tanimoto similarity to the training set, where similarity was computed using ECFP fingerprints. This result indicates that Vilya-1 is not simply memorizing and interpolating between structures seen during training - it has the ability to accurately sample conformers of molecules that differ substantially from its training set.

\paragraph{Vilya-1 generates chemically valid conformations.} Using the same set of held-out molecules, we assessed the chemical validity of the generated conformers with the PoseBusters toolkit and compared the results to those obtained for the corresponding reference X-ray structures deposited in the CSD \cite{buttenschoen2024posebusters}. As shown in Fig \ref{fig:fig3}B, conformers generated by Vilya-1 satisfy PoseBusters validity criteria at rates comparable to experimental structures. We observe an approximately 1\% occurrence of incorrect chirality at tetrahedral stereocenters; in practice, such conformers are easily identified and excluded from downstream analyses during inference. Double-bond stereochemistry is not explicitly encoded in the input representation, allowing the model to infer the preferred isomeric state from the context of the sampled three-dimensional structure. Despite the absence of such supervision, Vilya-1 still reproduces experimentally observed cis/trans configurations in 98\% of cases (green bar at the bottom of Fig \ref{fig:fig3}B).

\subsection{Vilya-1 generalizes to confidence estimation and property prediction}
\paragraph{Vilya-1 confidence module is comparably performant with machine learned interatomic potentials.} In the previous section, we evaluated the sampling efficiency of Vilya-1 and showed that, with 100 sampled conformers, the model captures biologically relevant conformations of macrocycles and small molecules far more frequently than existing approaches. For practical applications, however, it is desirable to further reduce the number of conformers passed to downstream pipelines by prioritizing the most energetically favorable structures. To address this need, we developed a confidence estimator trained to identify conformers that are most likely to resemble those observed in experiments. This strategy\textemdash central to modern protein structure prediction networks such as AlphaFold, where confidence metrics guide structure selection\textemdash has been widely validated for proteins but has not previously been established for macrocycle or small molecule conformer samplers. Our confidence module extends this paradigm, enabling principled and streamlined ranking of conformers in chemical spaces.

\begin{figure}[htbp]
  \centering
  \includegraphics[width=1.0\linewidth]{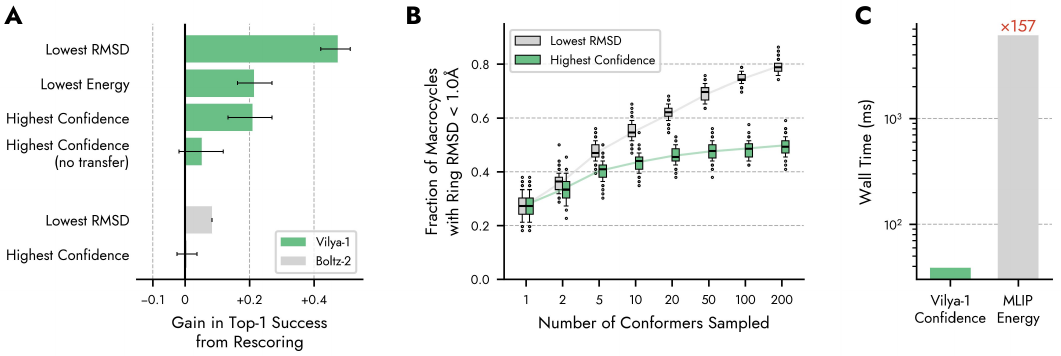}
  \caption{Conformer scoring with the Vilya-1 confidence model. A) Gain in top-1 success rates (ring RMSD $< 1$\AA) relative to random selection when conformers are ranked by the Vilya-1 confidence model, MLIP-based energy scoring, and Boltz-2, compared against the empirical upper bound (indicated by the ``Lowest RMSD’’). The ``no transfer’’ bar corresponds to a Vilya-1 confidence model trained from random initialization rather than fine-tuned from the pretrained conformer generator. B) Top-1 success rates of the Vilya-1 confidence model as a function of the number of sampled conformers. C) Computational efficiency comparison between Vilya-1 and MLIP scoring. Bars represent the average wall-clock time required to score a single conformer, calculated across 6,600 samples (100 conformers for each of 66 macrocycles). Average scoring time per conformer is 40ms and 6120ms for Vilya-1 and MLIP, respectively. }
  \label{fig:fig4}
\end{figure}

To benchmark the effectiveness of our confidence estimator, we performed a head-to-head comparison against a machine-learned interatomic potential (MLIP) \cite{fu2025learning}. For this analysis, we used the same sets of 100 conformers generated by Vilya-1, minimized each using AIMNet2 \cite{anstine2025aimnet2}, and computed single-point DFT-level energies with the eSEN Small Conserving model \cite{levine2025open} which served as MLIP-based scores. As in previous analyses, we assumed that the ground-truth crystal structure corresponds to an energy minimum\textemdash an approximation that is not strictly guaranteed, as removing a molecule from its crystal lattice or bound complex can alter its preferred states. Notably, AIMNet2 minimization produced only minor structural changes with mean pre- and post-min all atom RMSDs of 0.79\AA\ (Fig \ref{fig:figs4}), indicating that Vilya-1 routinely samples conformers already close to local energy minima. 

We quantified the scoring performance of each method by measuring the fraction of test-set molecules for which the top-ranked conformer was within 1.0\AA\ ring RMSD of the reference structure. As shown in Fig \ref{fig:fig4}A, the Vilya-1 confidence model increases the probability of selecting near-reference conformers by 21.1\% relative to a random selection, achieving performance comparable to MLIP-based scoring (21.9\%). Crucially, this performance was enabled by transfer learning from the conformer generator: when the confidence model is trained from random initialization (``no transfer’’), its scoring performance is not significantly different from random selection. Fig \ref{fig:fig4}B highlights the interplay between conformational sampling and confidence-based ranking: as the number of sampled conformers increases, selecting the most confident conformer progressively improves accuracy, rising from 27.6\% with a single sampled conformer to 48.7\% with 100 conformers. Despite similar accuracy, MLIP-based scoring is substantially more computationally expensive, requiring an additional minimization process. In contrast, the Vilya-1 confidence model adds negligible computational overhead (Fig \ref{fig:fig4}C).

As an additional baseline, we evaluated Boltz-2, using it both to generate and to score its own conformers. The combined sampling-and-scoring results shown in Fig \ref{fig:fig2}A indicate that, consistent with Fig \ref{fig:fig1}, Boltz-2 exhibits markedly lower sampling efficiency compared to Vilya-1 and ranking conformers by its confidence score provides no improvement over random selection. 

\paragraph{Vilya-1 enables high-resolution property prediction in real world macrocyclic drug discovery campaigns.} Estimating physico-chemical properties of macrocycles enables the evaluation and design of known drug-like features important for bioavailability and exposure. Two primary factors were evaluated in training and validating property prediction models: (1) model architecture and (2) initialization strategy (e.g. training from scratch versus fine-tuning from pretrained weights). Model architecture determines how effectively the model converts molecular inputs into informative representations and maps those representations to the target property. Initialization, in contrast, influences optimization efficiency by transferring patterns learned from a pretraining task that can accelerate\textemdash or, in some cases, hinder\textemdash learning on the downstream property task. Vilya-1’s structural performance demonstrates the value of its inductive biases at molecular understanding, and provides an opportunity to show the effectiveness of structure-based pretraining on predicting important molecular properties.

\begin{figure}[htbp]
  \centering
  \includegraphics[width=1.0\linewidth]{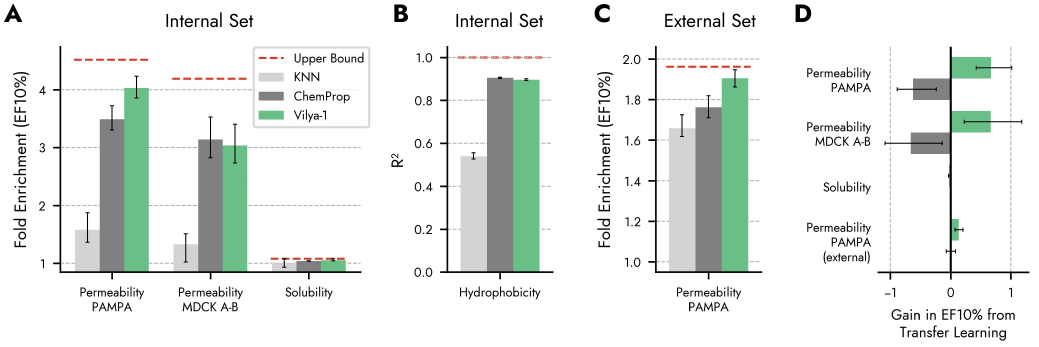}
  \caption{  Property prediction with Vilya-1: A) Enrichment factor (top 10\%) for property prediction on the internal dataset. All models are evaluated as ensembles of five, with error bars showing 95\% confidence intervals. Vilya-1 is initialized from the conformer generator. B) Performance on the hydrophobicity regression task. C) Enrichment factor for permeability prediction on the external benchmark. D) Effect of transfer learning on permeability and solubility prediction, showing changes in enrichment when models are fine-tuned from pretrained weights. Pretraining from Vilya-1’s conformer-generation task consistently improves performance on both internal and external datasets, whereas ChemProp pretraining from CheMeleon results in negative transfer on the internal set. }
  \label{fig:fig5}
\end{figure}

Vilya-1 is compared to popular property prediction baselines across different properties, as shown in Fig \ref{fig:fig5}. The largest gap in performance is seen on permeability (PAMPA) prediction on the internal set, which includes the most data on measurements across closely related analogs within the same chemical series. This is often the case in practical drug discovery workflows, where screened compounds share high chemical similarity as they originate from a related chemical series or common ring scaffold (Fig \ref{fig:figs6}). Predictive modeling in this regime demands higher ``resolution’’ feature extraction: conventional fingerprints or graph-based encodings may not sufficiently distinguish closely related analogs when subtle structural changes drive meaningful property shifts. This setting differs from many external datasets and published benchmarks, which are often more chemically diverse even when they share broad design constraints. This is reflected in comparison of Fig \ref{fig:fig5} panels A and C where resolving fine-grained analog differences in the internal permeability prediction is most critical and access to conformer information provides a clearer advantage and a larger gap in performance.

When Vilya-1’s property prediction model is initialized from its conformer-generation pretraining, performance is generally improved on both internal and external property tasks (Fig \ref{fig:fig5}D). By comparison, transfer learning from ChemProp’s pretrained model (CheMeleon) produces negative transfer in the internal properties. This divergence suggests that the representations learned during Vilya-1’s conformer-generation task can generalize more reliably and translate more effectively to feature extraction for downstream property prediction.

\subsection{Design and Screening}
Accurate structure and property prediction enable multiple important applications in drug discovery, including design and screening. In this section, we discuss using Vilya-1 to generate energy landscapes that are predictive of binding, to re-loop motifs derived from existing biologically-active peptides, and to synergize with commonly used display technologies.

\paragraph{Vilya-1 enables energy landscape analyses when coupled with MLIP.} \textit{De novo} macrocycle design often seeks sequences that predominantly adopt a single, stable conformation, thereby minimizing the entropic cost of binding under the assumption that the bound and solution structures are similar. The folding propensity of a sequence can be quantified using $P_{\textrm{near}}$ \cite{bhardwaj2016accurate}, which, given a conformational ensemble and an energy function, estimates the Boltzmann-weighted population of structures close to a binding-competent conformation. Values of $P_{\textrm{near}}$ range from 0, indicating negligible population of the target conformation, to 1, indicating strong preorganization.

In practice, the utility of $P_{\textrm{near}}$ is limited by conformational sampling. Biased sampling near the designed structure can overestimate preorganization, whereas insufficient sampling of distant conformations can underestimate it. Classical approaches mitigate these effects through extensive sampling or repeated energy minimization, often requiring tens to hundreds of thousands of conformers and limiting the applicability of $P_{\textrm{near}}$-based analyses. As a diffusion-based model, Vilya-1 alleviates these sampling pathologies by using partial diffusion to ensure balanced coverage of relevant regions of conformational space. Each conformer is scored using MLIP models with DFT-level accuracy and $P_{\textrm{near}}$ is computed from the resultant energy landscape \cite{anstine2025aimnet2, levine2025open}.

\begin{figure}[htpb]
  \centering
  \includegraphics[width=0.83333\linewidth]{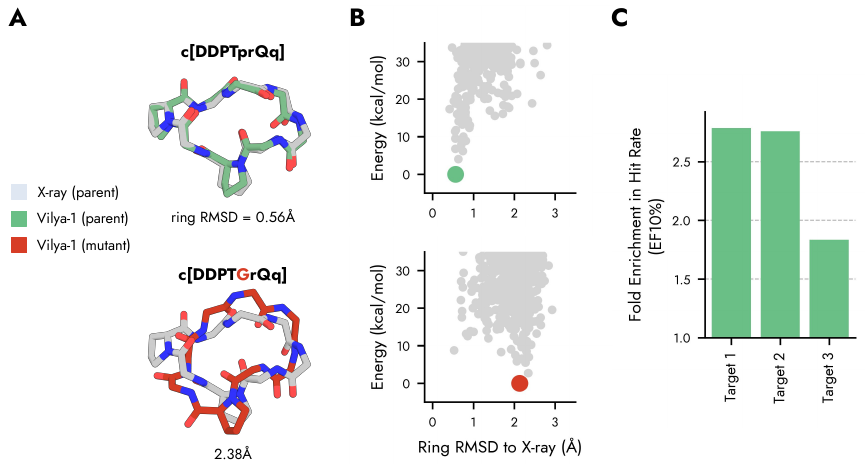}
  \caption{ A) Vilya-1-generated predictions and B) energy landscapes (upper row) of an experimentally validated heterochiral cyclic peptide (lower row, PDB ID 6BE7) \cite{hosseinzadeh2017comprehensive} and of a point mutant that replaces D-proline with glycine (lower row). This mutation results in a sequence that stabilizes an entirely different conformation than that adopted by the parent. Side-chains other than proline or \textsc{D}-proline are hidden for visual clarity. C) Retrospective fold enrichment in hit rate (EF10\%) for previously designed macrocycles, when samples are prioritized based on $P_{\textrm{near}}$ relative to the intended designed structure. }
  \label{fig:fig6}
\end{figure}

Figure \ref{fig:fig6}A and B showcase the ability of this pipeline to detect the impact that small modifications in chemical structures have on the conformational ensemble sampled by a previously described \textit{de novo} designed macrocycle \cite{hosseinzadeh2017comprehensive}. The energy landscape sampled by the parent design (Fig \ref{fig:fig6}A) shows a funnel-like quality: conformations closer to the designed structure are lower in energy than conformations with higher ring RMSD. The $P_{\textrm{near}}$ of this landscape is $\sim$1.0, indicating that this design very likely adopts the designed conformation in solution. Mutating a \textsc{D}-proline to a glycine in Fig \ref{fig:fig6}B ablates this funnel-like quality and instead results in a sequence that favors an entirely different conformation. This is apparent in the $P_{\textrm{near}}$ of the mutant sequence, which is close to 0, and showcases how $P_{\textrm{near}}$ can be used to detect how single point mutations may alter the conformational preferences of macrocyclic peptides.

We ranked compounds using $P_{\textrm{near}}$ calculated from Vilya-1-generated ensembles across three independent drug discovery campaigns with experimentally measured binding affinities. These de-novo-designed peptide macrocycles all result from either previously described or proprietary-to-Vilya design processes. As such, they represent a high-quality, pre-filtered set that has undergone multiple rounds of filtering and prioritization prior to synthesis. Across these independent targets, a macrocycle is considered a hit if its Kd $\leq 50 \mu\textrm{M}$. In all three campaigns, $P_{\textrm{near}}$ calculated from Vilya-1-generated ensembles enriched for binders (Fig \ref{fig:fig6}C). This indicates that macrocycles pre-organized into their binding-competent conformations are more likely to bind their target proteins, and that accurate conformational sampling facilitates the discovery of such macrocycles. Notably, this enrichment is target-independent: our model identifies potential binders solely on ligand conformational landscapes, without explicitly modeling interactions with the target protein or pretraining on this task.

\begin{figure}[htpb]
  \centering
  \includegraphics[width=1.0\linewidth]{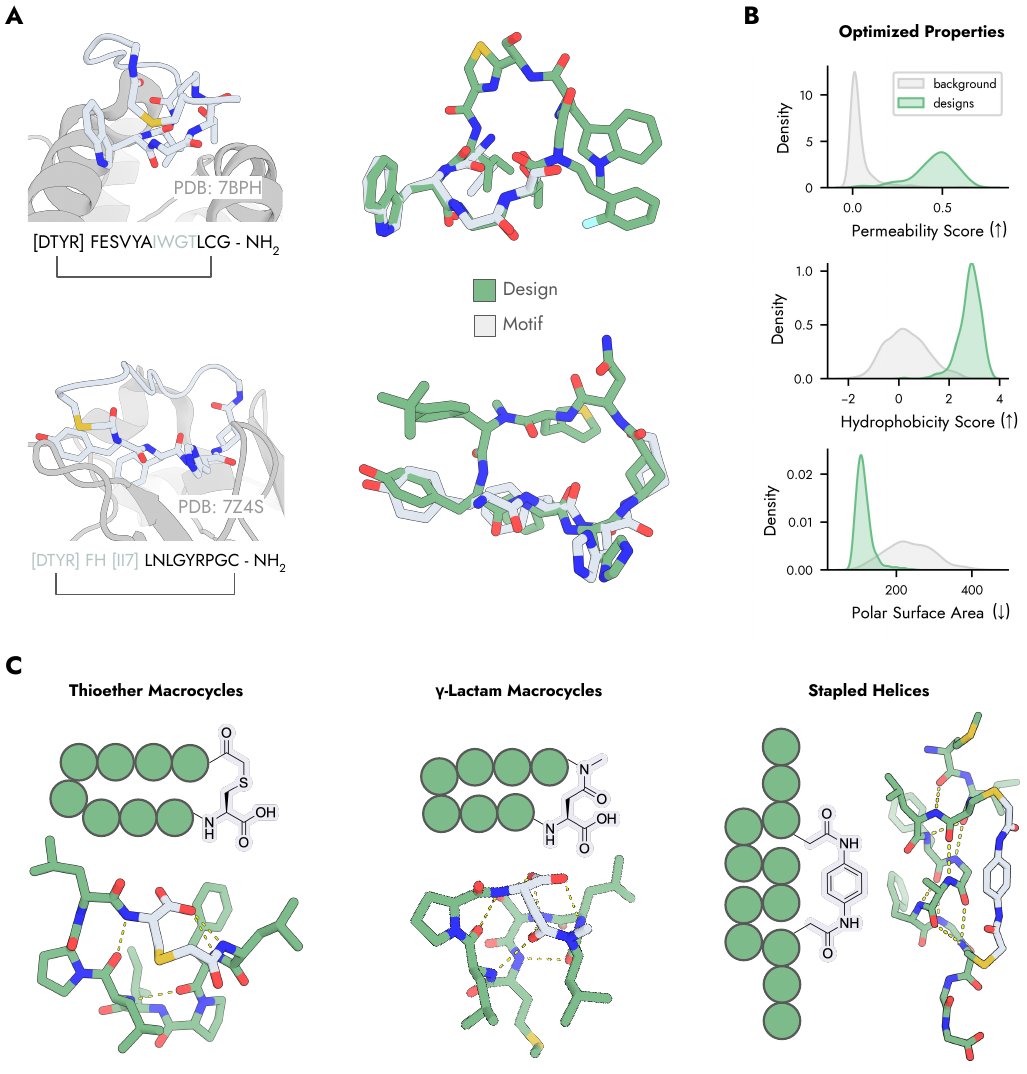}
  \caption{ Computational design and optimization of peptides using non-canonical amino acids and linkages. A) Two thioether macrocycle binders identified from mRNA display, the residues highlighted in green make up the majority of the interface. Using our design methodology, Vilya-1 can be used to miniaturize large peptides using non-canonical amino acids, while maintaining the conformation of the motif. B) From the randomly-initialized sequences, the loss function successfully optimizes key properties for drug-like macrocycles. C) Vilya-1 can be used to design with cyclization chemistries commonly used in library and display technologies. }
  \label{fig:fig7}
\end{figure}

\paragraph{Vilya-1 enables sequence optimization biased by structure- and property-based constraints.} Optimizing macrocycles towards drug-likeness often involves maintaining the pre-organization of key binding moieties in 3D while removing portions of the molecule to drive towards particular properties. The binding moieties may be identified from SAR experiments of a parent hit or extracted from high-resolution structures of the target of interest. Such motif scaffolding-based design approaches are common in molecular design. Coupling Vilya-1 structure and property prediction modules enables identification of novel sequences capable of maintaining the preorganization of the motif while improving important properties \textit{in silico}. We employ Vilya-1 in a discrete, multi-objective optimization procedure to generate candidate sequences for a range of design problems. Unlike many existing design approaches, this process is general and not limited to head-to-tail chemistries or canonical residues.

In Fig \ref{fig:fig7}A, we showcase our design process miniaturizing a 13- and a 14-residue macrocycle identified from display-based selection methods targeting intracellular proteins \cite{dai2022state, miura2023vitro}. In particular, the macrocycle selected to bind the main protease from SARS-CoV-2 (Fig \ref{fig:fig7}A, bottom) was identified from a library specifically designed to contain $\gamma$-amino acids. We use Vilya-1 to miniaturize these peptides by excising the key binding motif from the parent macrocycle and using our pipeline to insert new building blocks to hold the motif in the binding competent conformation. Both output designs are 7 residues long, and Vilya-1 successfully models non-canonical residues in both the motif and in the relooped scaffold. Example designs are depicted superimposed to their respective motifs.

In addition to preserving the motif, we optimize for important drug-like properties. Comparing initially-sampled sequences to those identified at the end of the design process, we find the sequences have been optimized towards higher predicted permeability, higher predicted hydrophobicity, and lower solvent-accessible polar surface area (Fig \ref{fig:fig7}B).

\paragraph{Vilya-1 enables design of macrocycles with complex topologies.} Peptidic macrocycles are a broad class of molecule. They are constructed from amino acid building blocks\textemdash both canonical and non-canonical\textemdash that can be connected in myriad topologies \cite{hayes2021approaches}. The development and application of computational design methods to peptide macrocycles have largely been focused on head-to-tail cyclic peptides, in which the N- and C- terminal amino acids are connected by an amide bond. This is in part because many such methods repurpose large protein structure prediction networks to design and model macrocycles \cite{hosseinzadeh2017comprehensive, rettie2025cyclic, rettie2025accurate}. In contrast, high-throughput experimental methods often employ non-head-to-tail topologies due to nuances in the underlying display technology \cite{josephson2014mrna, huang2018rna, quartararo2020ultra, goto2021rapid, plais2022macrocyclic}. As a consequence, many computational design tools\textemdash deep learning-based or otherwise\textemdash struggle to synergize with the various display-based screening technologies that have previously dominated macrocycle drug discovery campaigns.

In contrast, the all-atom representation employed by Vilya-1 is agnostic to macrocycle topology. This allows the discrete optimization procedure to explore scaffold architectures commonly accessed by high-throughput display technologies, including thioether-based and $\gamma$-lactam linkages seen in mRNA display \cite{goto2021rapid, ohta2023validation} and $i,i+7$ helix staples present in stapled helix-based phage display platforms \cite{li2022novo}. We depict example designs for all three topologies in Fig \ref{fig:fig7}C. These exotic linkages are inaccessible or difficult to model by many existing tools, particularly those designed to model natural proteins. 

\section{Conclusion}
We introduce Vilya-1, a state-of-the-art deep learning model for macrocycle conformer generation and design. Powered by a novel all-atom architecture, Vilya-1 substantially improves sampling of experimentally observed macrocycle structures across benchmarks spanning X-ray, NMR, and receptor-bound conformations, outperforming physics-based protocols, small-molecule conformer generators, and co-folding networks. The gains are especially pronounced for non-canonical chemistries and mixed peptidic-non-peptidic scaffolds. Importantly, this performance is not restricted to cyclic peptides or molecules close to the training distribution: Vilya-1 also generalizes to small molecules and produces conformers that satisfy standard chemical validity criteria at rates comparable to experimental structures.

Beyond structure sampling, Vilya-1 serves as a foundation model that can be fine-tuned for downstream tasks relevant to macrocycle drug discovery. First, its confidence module enables principled ranking within generated ensembles, offering a lightweight alternative to energy-based rescoring that approaches the performance of MLIP methods. Second, structure-based pretraining transfers to property prediction, improving prioritization of developability-relevant properties such as permeability and hydrophobicity, with the largest gains in the hit-series regime where closely related analogs must be distinguished by subtle, structure-dependent effects. Third, Vilya-1 supports design workflows that reflect real macrocycle campaigns, including ligand-centric conformational landscape analyses for binding enrichment and discrete optimization that preserves binding motifs while adapting scaffold size, chemistry and properties. Because the model operates on a unified all-atom representation, these capabilities extend naturally to diverse cyclization chemistries and topologies common in display technologies, enabling design beyond canonical head-to-tail scaffolds.

Together, these results position Vilya-1 as a general platform for macrocycle modeling, linking structure sampling, confidence estimation, property prediction, and topology-aware design within a single framework. We anticipate Vilya-1 will accelerate macrocycle discovery by reducing reliance on chemistry-specific sampling protocols, enabling higher-resolution prioritization within medicinal chemistry series, and expanding the range of macrocycle topologies that can be explored for therapeutic development.

\section{Contributors}

Pascal Sturmfels, Milad Salem, Naozumi Hiranuma, Stephen Rettie, Xiaoliang Pan, Benjamin D. Sellers, Adam P. Moyer, Patrick J. Salveson*, Ivan Anishchanka*

*Correspondance to: \href{mailto:patrick@vilyatx.com}{patrick@vilyatx.com}, \href{mailto:ivan@vilyatx.com}{ivan@vilyatx.com}.

\printbibliography

\appendix

\setcounter{figure}{0}
\renewcommand{\thefigure}{S\arabic{figure}}

\setcounter{table}{0}
\renewcommand{\thetable}{S\arabic{table}}

\section{Supplement}

\subsection{Macrocycle validation set} AAGAGG10, ABENEC, ADUGAH, ADUHUC, AFOKUE, AFUWON, AMEHUX, ANUJUQ, ANUTOU, AXAWAZ, BAPLUD, BAXBUB, BAXPEX, BICMEF, BIDPUC, BIDQEN, BIDYIZ, BOPZDT10, BUTZEY, CAHWEN, CALSAR, CAMVES11, CAZVEF, CAZZEK, CETKAQ, CETSAW, CIYMAB, COGGIO, CUQYUI, DACMAW, DALPUC, DICWET, DOVXUI, DOVYAP, DOXKEI, EPOZOA, EVEDEO, FAGFEZ, FALLIO, FAQYEC, FAWYEI, FEXLAW, FIFSOE, FIGZAW, FIXYAM, GAFSEL, GLSARM, GOCTUO, GOVNOU, GUJNOQ, HABTIN, HEBHUR, HEBLIJ, HECHIG, HEGNOY, HEHKUE, HEWKEA, HEYZAN, HIXHEE, IBADUO, IBAGAX, IQELAU, IQIFIY, IQIFOE, ITABAJ, IYONIU, IYURAV, JEPFES, JEPFIW, JEXDAU, JINWUZ, JUHPUX, KAJJOX, KEFWUN, KEYBOH, KORMEJ, KUFNOR, KUFPUW, LALWIF, LEQRUW, LEYCAV, LINTUZ, LOLMOQ, LOXSOJ, LUXHUH, MADKIN, MAZMCP, MEANLD, MOSDOO01, MOSDUU, MOSFAC, MOSFEG, NAFJEJ, NANRAW, NAWLON, NAWTUC, NIVQOZ, NIYQUK, NOJZER, NUGTAL, ODETEW, ODETIA, ODULAB, OGAZUS, OGUQEN, OHIFAM, OJIDIU, OLAMIZ, OVUZAJ, PEJXIN, PERPOT, PEZHEK, PEZNAJ, PILXAK, PIZYIH, QANGER, QARJIG, QAXWAQ, QAZZID, QEDJOA, QENREG, QIJJAX, QIRRUE, QOQXEB, QOSNAN, QUMBUV, QUWMOM, RAQVOU, ROXLEX, RUMKIT01, RUNDOU, RUNPOF, RUTLEY, RUVJOK01, SAVHUV, SAVLOS, SAVLUY, SAZNUE, SBAGMA10, SIKMUU10, SUJMOZ, SUPLEU, TADMIX, TAFROL, TALZUC, TIDDAL, TIKDUO, TOZBER, TUGYEZ, TZCTDO10, UCASUP, UCUDEC, UGENUS, UKATUY, UKAVOU, UKOYAU, UKUVEB, ULIPIO, UNONES, UQAMUX, URUQAA, UXOHIZ, VAFPEX, VAHGOA, VAMNIJ, VAMNOP, VECFIT, VECKUJ, VEHGEX, VOLHIO, VOLHOU, VOLKIQ, VONKOY, VORZEH, VUHHEM, WAFHER, WASFED, WEMGON, WEMGUT, WENFOL, WETYAW, WIVGUG, WIWXOQ, WUSBET, WUWDOJ, XACJAM, XAMYOB, XEFLIE, XEVGUD, XIWDUC, YACJAP, YAGVEH, YEHSOV, YOCXET, YOJNUH, YOXTAI, YUPHEW, ZAMNIK, ZEHMOR, ZEWREA, ZILQOA01, ZOZFOL

\subsection{Macrocycle NMR test set}

\paragraph{Canonical.} 1BRV\_A, 1EDP\_A, 1EGT\_A, 1FGD\_A, 1FGE\_A, 1FUL\_A, 1FUV\_A, 1GM2\_A, 1GNA\_A, 1GNB\_A, 1HVZ\_A, 1IM1\_A, 1JBL\_A, 1JBN\_A, 1KWD\_A, 1KWE\_A, 1LB7\_A, 1M4E\_A, 1MA2\_A, 1MA5\_A, 1N2Y\_A, 1PMX\_B, 1QVK\_A, 1QVL\_A, 1R8T\_A, 1SKI\_A, 1SKK\_A, 1TMR\_A, 1UYA\_A, 1UYB\_A, 1V46\_A, 1X7K\_A, 1Y49\_A, 1YT6\_A, 2A9X\_B, 2ATG\_A, 2BEY\_A, 2EFZ\_A, 2G6U\_A, 2IFJ\_A, 2IGU\_A, 2JNR\_A, 2JQW\_A, 2JUQ\_A, 2JUR\_A, 2JUS\_A, 2JUT\_A, 2L07\_A, 2LER\_A, 2LS1\_A, 2LU6\_A, 2LWB\_A, 2LWQ\_A, 2LWS\_A, 2LWT\_A, 2LWU\_A, 2LWV\_A, 2LYE\_A, 2LYF\_A, 2LZI\_A, 2M2Y\_A, 2M3N\_A, 2M77\_A, 2M78\_A, 2M79\_A, 2MD6\_A, 2MGO\_A, 2MI1\_A, 2MTM\_A, 2MUH\_A, 2NB5\_A, 2NB6\_A, 2NC7\_A, 2NDL\_A, 2NDM\_A, 2NDN\_A, 2NS4\_A, 2OTQ\_A, 2OX2\_A, 2P7R\_A, 5H1H\_A, 5H1I\_A, 5JPL\_A, 5LFF\_A, 5VAV\_A, 5XIV\_A, 6BX9\_A, 6DNY\_A, 6EY3\_A, 6FGM\_A, 6NOX\_A, 6OQP\_A, 6OTA\_A, 6OTB\_A, 6OVJ\_A, 6PIN\_A, 6PIO\_A, 6PIP\_A, 6QKF\_A, 6U7Q\_A, 6U7R\_A, 6U7S\_A, 6WPV\_A, 6XTH\_A, 6XTI\_A, 7ALD\_A, 7CSS\_A, 7CU6\_A, 7EDK\_A, 7F32\_A, 7K1M\_A, 7L7A\_A, 7LL7\_B, 7LQS\_A, 7M25\_A, 7M27\_A, 7M28\_A, 7M29\_A, 7M2A\_A, 7M2B\_A, 7M2C\_A, 7N0T\_A, 7N23\_A, 7QLF\_A, 7SAV\_A, 7SAW\_A, 7V5E\_A, 8B4R\_A, 8F04\_A, 8IKQ\_A, 8IL1\_A, 8IL2\_A, 8IL6\_A, 8K3M\_A, 8K3N\_A, 8XTH\_A, 9BHN\_A

\paragraph{Non-canonical.} 185D\_A, 193D\_C, 1A1P\_A, 1B45\_A, 1C4B\_A, 1CNL\_A, 1CSA\_A, 1CYA\_A, 1CYB\_A, 1D7T\_A, 1DFY\_A, 1DFZ\_A, 1DG0\_A, 1DG2\_A, 1E74\_A, 1E75\_A, 1E76\_A, 1EVB\_A, 1FOZ\_A, 1G2G\_A, 1GJE\_A, 1HD9\_A, 1I6Y\_A, 1I8E\_A, 1I93\_A, 1I98\_A, 1IEN\_A, 1IEO\_A, 1IM7\_A, 1IMI\_A, 1IMW\_A, 1J8N\_A, 1JBF\_A, 1JD8\_A, 1K64\_A, 1KB7\_A, 1KB8\_A, 1KFP\_A, 1KVF\_A, 1KVG\_A, 1LCM\_A, 1M2C\_A, 1MII\_A, 1MPV\_A, 1MTQ\_A, 1MXN\_A, 1MXP\_A, 1N09\_A, 1N0A\_A, 1N0C\_A, 1N0D\_A, 1NIL\_A, 1NIM\_A, 1PAJ\_A, 1PAK\_A, 1PAN\_A, 1PAO\_A, 1PG1\_A, 1QFB\_A, 1QMW\_A, 1QS3\_A, 1QX9\_A, 1RGR\_B, 1RKK\_A, 1SKL\_A, 1SOC\_A, 1T9E\_A, 1UL2\_A, 1WCT\_A and B, 1WO0\_A, 1WO1\_A, 1XBH\_A, 1XGA\_A, 1XGB\_A, 1XGC\_A, 1XXZ\_A, 1XY4\_A, 1XY5\_A, 1XY6\_A, 1XY8\_A, 1XY9\_A, 1YL8\_A, 1YL9\_A, 1ZLC\_A, 2B5K\_A, 2B5P\_A, 2B5Q\_A, 2EW4\_A, 2FR9\_A, 2FRB\_A, 2GCZ\_A, 2H8S\_A, 2I28\_A, 2IFI\_A, 2IFZ\_A, 2IGZ\_A, 2IH0\_A, 2IH6\_A, 2IH7\_A, 2KDQ\_A, 2KX5\_B, 2LR9\_A, 2LXG\_A, 2LZ5\_A, 2M1P\_A, 2M2G\_A, 2M2H\_A, 2M2S\_A, 2M2X\_A, 2M3I\_A, 2M61\_A, 2M62\_A, 2M6C\_A, 2M6D\_A, 2M6E\_A, 2M6F\_A, 2M6G\_A, 2M6H\_A, 2M7I\_A, 2M7J\_A, 2MDB\_A, 2MDQ\_A, 2MFX\_A, 2MFY\_A, 2MG6\_A, 2MOA\_A, 2MTO\_A, 2MTT\_A, 2MTU\_A, 2NAY\_A, 2NBC\_A, 2NS3\_A, 2RTV\_A, 2RU2\_A, 2SOC\_A, 3CYS\_B, 3ZKT\_A, 5KX2\_A, 5L34\_A, 5LFH\_A, 5OJT\_A, 5OLF\_A, 5TWW\_A, 5UG3\_A, 5UG5\_A, 5VR1\_A, 6AZA\_A, 6B34\_A, 6B35\_A, 6BE7\_A, 6BE9\_A, 6BEN\_A, 6BEO\_A, 6BER\_A, 6BES\_A, 6BET\_A, 6BEU\_A, 6BEW\_A, 6BF3\_A, 6BF5\_A, 6BVU\_A, 6BVW\_A, 6BVX\_A, 6BVY\_A, 6D2U\_A, 6HVB\_A, 6HVC\_A, 6IMG\_A, 6IMH\_A, 6KMY\_A, 6KN2\_A, 6KN3\_A, 6KNO\_A, 6KNP\_A, 6MY1\_A, 6MY2\_A, 6MY3\_A, 6OC2\_A, 6PI2\_A, 6PI3\_A, 6R28\_A, 6U24\_A, 6U7U\_A, 6U7W\_A, 6U7X\_A, 6UXS\_A, 6V4I\_A, 6VHJ\_A, 6VLJ\_A, 6VY8\_A, 6Y1Q\_A, 7D37\_A, 7JU9\_A, 7KNN\_A, 7L96\_A, 7L98\_A, 7L9D\_A, 7M7X\_A, 7N22\_A, 7N26\_A, 7OFG\_A, 7OXF\_A, 7QS6\_B, 7TVQ\_A, 7TVR\_A, 7TZ3\_A, 7UBC\_A, 7UBD\_A, 7UBE\_A, 7UBF\_A, 7UBG\_A, 7UBH\_A, 7UBI\_A, 7UZL\_A, 7ZAX\_B, 7ZED\_B, 8BSS\_B, 8CTO\_A, 8CUN\_A, 8CWA\_A, 8FLP\_A, 8HR3\_A, 8HR4\_A, 8ONU\_B, 8Q7J\_Z, 8QAQ\_Z, 8QAS\_Z, 8QBP\_Z, 8W16\_A, 8W17\_A, 8X3N\_A, 8X40\_A, 9BAF\_A, 9BFL\_A, 9EH4\_A

\newpage
\subsection{Bioactive Macrocycles}

\begin{table}[htbp]
\centering
\caption{PDB IDs for macrocycles shown in Fig 1A}
\begin{tabular}{ll}
\hline
Compound name & PDB ID \\
\hline
Lorlatinib & 4CLI \\
Griselimycin & 8CIX \\
$\alpha$-Amanitin & 2VUM \\
$\alpha$-Conotoxin & 1PEN \\
Octreotide & 6VC1 \\
FK506 & 1BFK \\
Sanglifehrin A & 1YND \\
Danoprevir & 3SU0 \\
Darobactin & 7NRF \\
SFTI-1 & 3P8F \\

\hline
\end{tabular}
\label{tab:tabs1}
\end{table}

\subsection{Supplemental Figures}

\begin{figure}[htbp]
  \centering
  \includegraphics[width=0.8\linewidth]{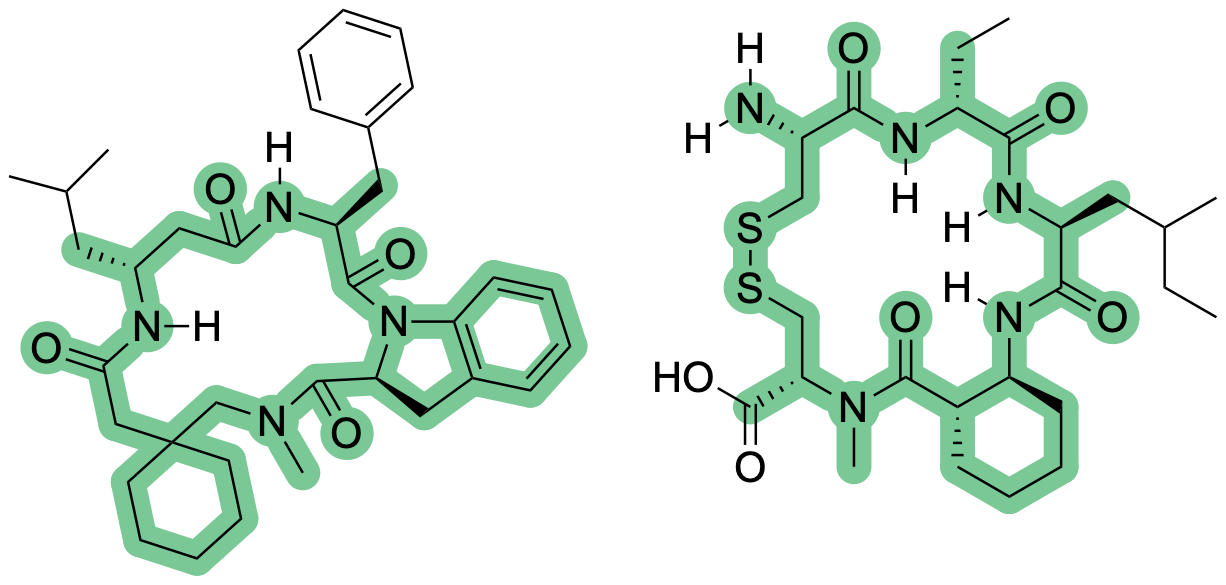}
  \caption{ Ring Scaffold definition. Example macrocyclic structures where atoms comprising the ring scaffold are highlighted in green. The highlighted set of atoms are used to assess RMSD between ground-truth and predicted structures of macrocycles. }
  \label{fig:figs1}
\end{figure}

\begin{figure}[htbp]
  \centering
  \includegraphics[width=0.5\linewidth]{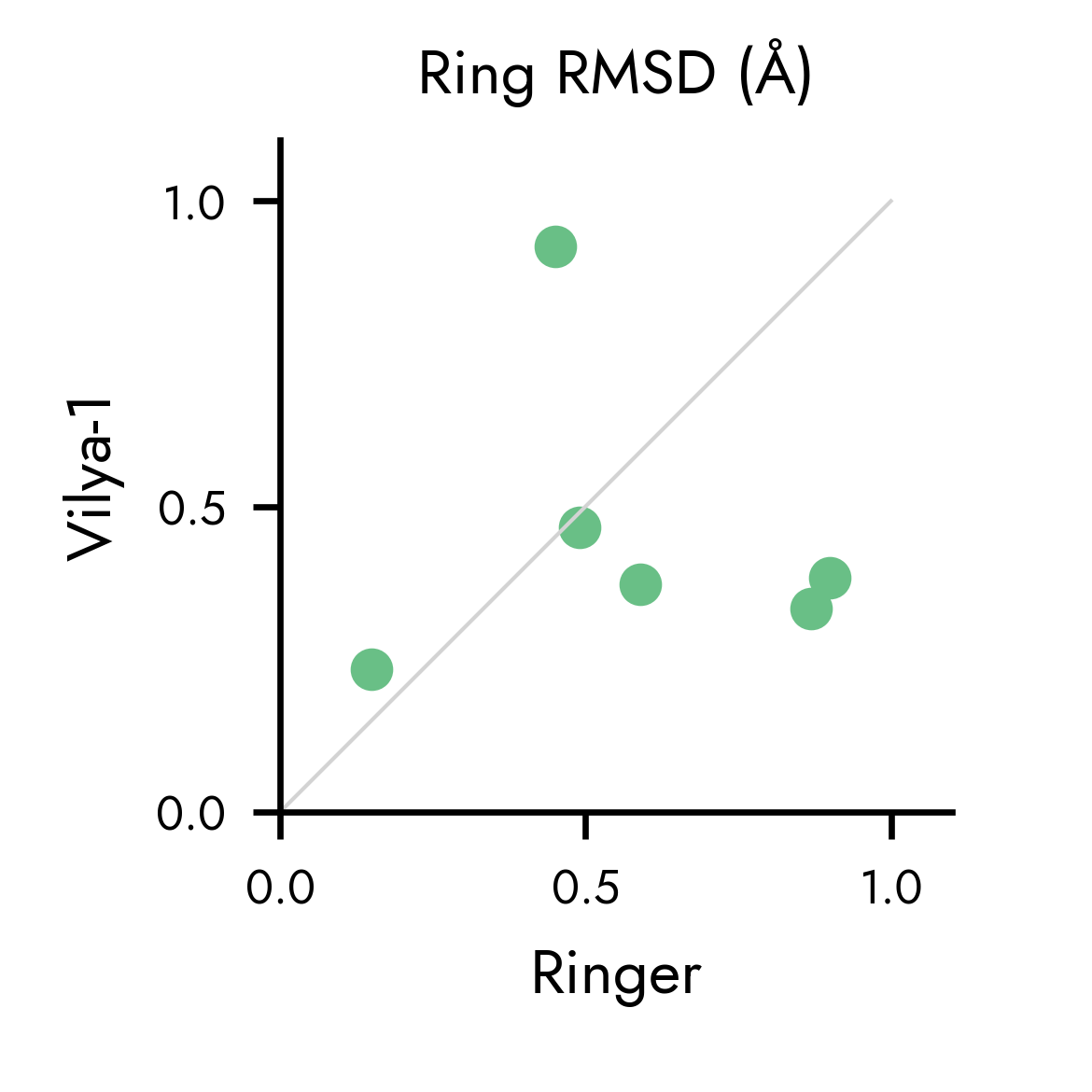}
  \caption{ Head-to-head comparison of Ringer and Vilya-1 of the 6 (out of 260) macrocycles in our PDB-derived test set that are fully canonical, head-to-tail cyclized, and have 6 or fewer residues. These constraints are imposed by the RINGER model. Vilya-1 outperforms RINGER on 4/6 molecules, but the difference between the two methods is not statistically significant given the small sample size. }
  \label{fig:figs2}
\end{figure}

\begin{figure}[htbp]
  \centering
  \includegraphics[width=0.9\linewidth]{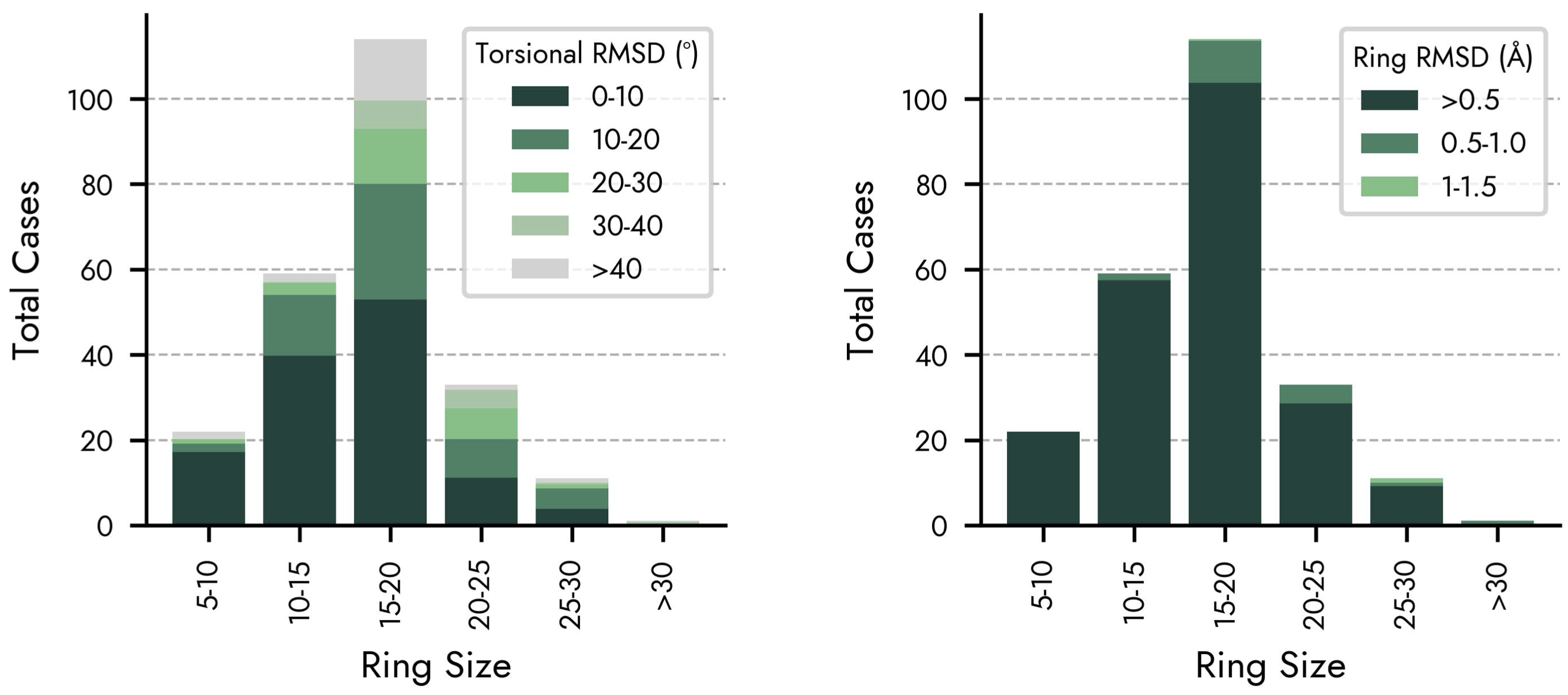}
  \caption{ Vilya-1 sampling accuracy of isolated ring systems. We reproduce the evaluation protocol of Fig \ref{fig:fig2} in Ref. \cite{robson2025accurate} to enable a direct, head-to-head comparison with Prime MCS. Results are based on ensembles of 1,000 sampled conformers per molecule, using the same ring definitions as in Ref. \cite{robson2025accurate}. }
  \label{fig:figs3}
\end{figure}

\begin{figure}[htbp]
  \centering
  \includegraphics[width=0.6\linewidth]{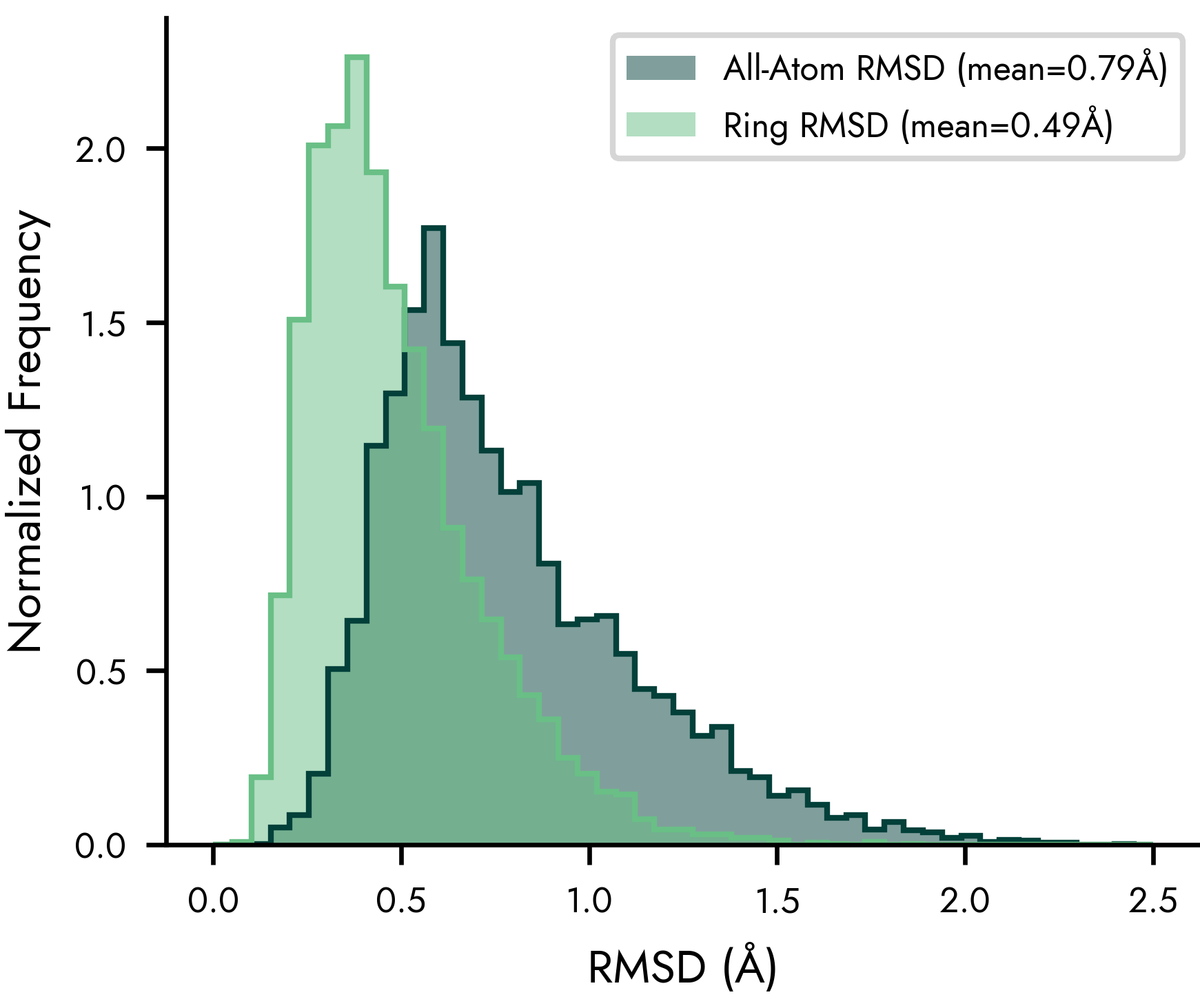}
  \caption{ All-atom and ring RMSD between conformers generated by Vilya-1 before and after AIMNet2 energy minimization. A total of 6,600 conformers (66 molecules, 100 conformers per molecule) were analyzed. The mean RMSD between pre- and post-minimization structures is 0.79\AA\ for all atoms and 0.49\AA\ for ring atoms. }
  \label{fig:figs4}
\end{figure}

\begin{figure}[htbp]
  \centering
  \includegraphics[width=0.4\linewidth]{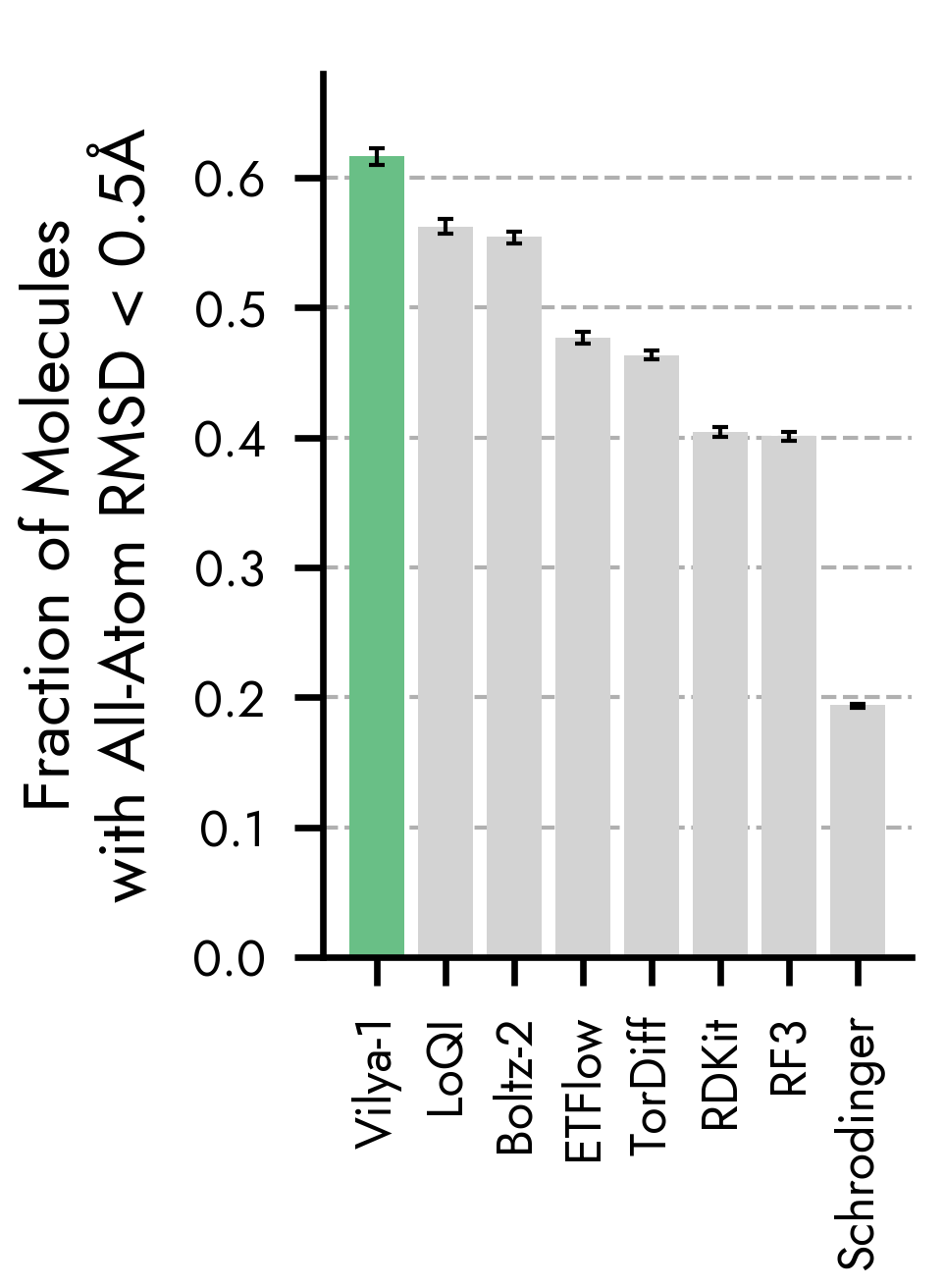}
  \caption{ Benchmarking results for Vilya-1 and alternative methods on the Platinum Diverse set of receptor-bound conformations of small molecules. }
  \label{fig:figs5}
\end{figure}

\begin{figure}[htbp]
  \centering
  \includegraphics[width=1.0\linewidth]{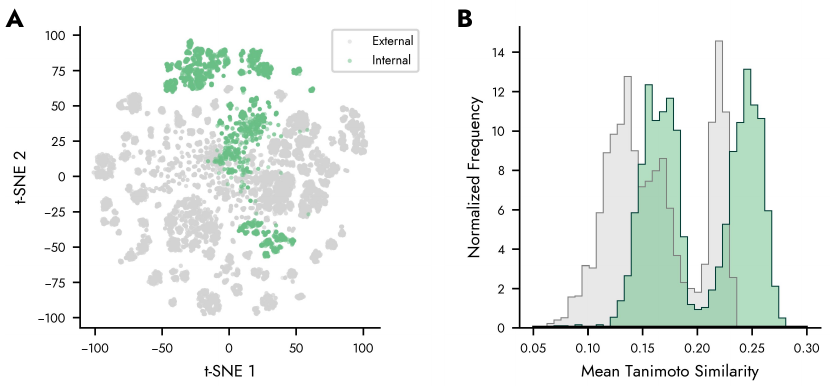}
  \caption{ A) tSNE plot of ECFP fingerprints on internal and external property sets. B) Mean Tanimoto similarity between the compounds of each set. }
  \label{fig:figs6}
\end{figure}

\begin{figure}[htbp]
  \centering
  \includegraphics[width=1.0\linewidth]{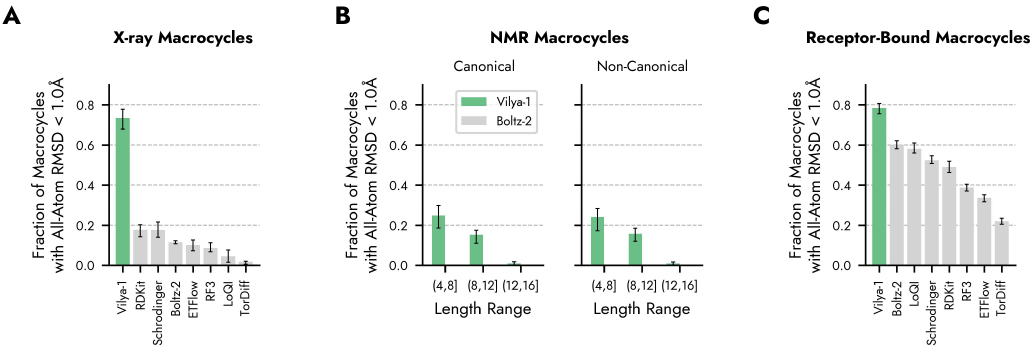}
  \caption{ Sampling efficiency of Vilya-1 measured by all-atom RMSD. Performance is shown for A) X-ray structures of cyclic peptides; B) solution NMR structures of canonical and non-canonical cyclic peptides; and C) receptor-bound conformations of cyclic peptides. Panels A--C correspond to Fig \ref{fig:fig1}B--D, respectively. The lower success rates observed for NMR structures in panel B when evaluated by all-atom RMSD likely reflect increased side-chain conformational heterogeneity and weaker experimental restraints in solution NMR ensembles compared to crystallographic structures. }
  \label{fig:figs7}
\end{figure}

\end{document}